\documentclass{article}

\usepackage{microtype}
\usepackage{graphicx}
\usepackage{subcaption}
\usepackage{booktabs}

\usepackage{hyperref}

\usepackage[accepted]{icml2026}

\usepackage{amsmath}
\usepackage{amssymb}
\usepackage{mathtools}
\usepackage{amsthm}

\usepackage[capitalize,noabbrev]{cleveref}

\theoremstyle{plain}

\theoremstyle{definition}

\theoremstyle{remark}

\usepackage[disable,textsize=tiny]{todonotes}

\icmltitlerunning{Tabular Population Priors for EHR Medication Set Prediction}

\begin{document}

\twocolumn[
  \icmltitle{Tabular Population Priors as Structured Retrieval for EHR Medication Set Prediction}

  \begin{icmlauthorlist}
    \icmlauthor{Animesh Agarwal}{optum}
    \icmlauthor{Meysam Ghaffari}{optum}
    \icmlauthor{Nina Fatehi}{optum}
    \icmlauthor{Carlos Morato}{optum}
  \end{icmlauthorlist}

  \icmlaffiliation{optum}{Optum AI, UnitedHealth Group, Minneapolis, Minnesota, USA}

  \icmlcorrespondingauthor{Animesh Agarwal}{animesh.agarwal@optum.com}

  \icmlkeywords{structured data, electronic health records, medication recommendation, clinical retrieval augmented generation, population priors}

  \vskip 0.3in
]

\printAffiliationsAndNotice{}

\begin{abstract}
Medication prediction from electronic health records is a fixed-vocabulary
structured prediction problem, yet recent clinical RAG systems often treat it
like open-ended case-based reasoning: retrieve similar patient admissions and
ask an LLM to infer a medication set. This retrieval object is indirect---the
model receives retrieved patient records while the target is a discrete set of
medication-class labels. We propose a more task-aligned retrieval object: a
tabular population prior---ranked medication-class candidates from a sparse
condition--medication co-occurrence table estimated from training admissions.
Using PACE-RAG as a fixed downstream pipeline and replacing only its patient
retriever, this drop-in swap improves F1 from 0.303 to 0.338 (+11.5\%
relative) and Jaccard from 0.218 to 0.237 on 1{,}004 held-out MIMIC-IV
admissions, with no additional LLM calls and no inference-time patient index.
A non-LLM top-$k$ baseline reaches F1=0.248, confirming the prior is
predictive before LLM integration. Aggregate population statistics can provide
a more task-aligned, efficient, and auditable retrieval source than
patient-level nearest neighbors for fixed-vocabulary EHR set prediction.
\end{abstract}
\section{Introduction}
\label{sec:intro}

Electronic health records (EHRs) are a canonical structured-data domain:
diagnoses, medications, procedures, laboratory events, and admissions are stored
as longitudinal tables \citep{johnson2023mimic}. Medication recommendation from
EHRs is therefore naturally framed as fixed-vocabulary set prediction: given the
current admission and available patient history, a system predicts a subset of
medication classes from a predefined vocabulary. Unlike open-ended clinical text
generation, the inputs are structured, the outputs are discrete labels, and
performance is measured by set overlap with observed medication classes.

Medication recommendation has traditionally been studied using task-specific
multi-label predictors over longitudinal patient records. Early neural methods
such as LEAP model the drug set sequentially from the current patient state
\citep{zhang2017leap}, while later approaches incorporate external memory,
graph structure, drug--drug interaction constraints, and molecular information
to improve recommendation quality and safety
\citep{shang2019gamenet,yang2021safedrug,wu2022cognet,yang2023molerec}.
Recent work also explores retrieval or gating over medication history for drug
recommendation \citep{meng2024mgrn}. These models directly learn a mapping
from patient histories to medication sets.

Clinical retrieval-augmented generation (RAG) systems take a different route:
they retrieve evidence and supply it to a language model for grounded
prediction. RAG was originally developed for knowledge-intensive NLP tasks
\citep{lewis2020rag}, often using dense retrieval and vector search
\citep{karpukhin2020dpr,johnson2019faiss}; clinical variants retrieve
guidelines, clinical knowledge, patient records, or treatment-relevant evidence
for clinical question answering, medical reasoning, and personalized treatment
recommendation \citep{zakka2023almanac,zhao2025medrag,liu2025treatrag}.
PACE-RAG is the most directly relevant example for our setting: it retrieves
similar patient admissions using FAISS and passes those admissions through a
multi-stage LLM pipeline for medication recommendation \citep{huh2026pacerag}.
This design raises a structured-data question that has received less attention:
for fixed-vocabulary EHR set prediction, is the most useful retrieval object a
small set of individual nearest-neighbor patients, or an aggregate summary of
population-level prescribing patterns?

We study this question on admission-level medication-class prediction from
MIMIC-IV \citep{johnson2023mimic}. Instead of indexing patient records, we
retrieve from a compact \emph{structured population prior}: a training-set
condition--medication table that maps current diagnoses to ranked
medication-class candidates. We use these candidates in place of
FAISS-retrieved patient records in PACE-RAG, while keeping the LLM backbone,
downstream stages, medication vocabulary, output parsing, and evaluation fixed.
This changes the evidence given to the LLM: individual nearest-neighbor
admissions are replaced by aggregate condition--medication statistics aligned
with the fixed output vocabulary. On 1{,}004 held-out MIMIC-IV admissions, the population-prior swap improves
PACE-RAG F1 from 0.303 to 0.338 and Jaccard from 0.218 to 0.237. A non-LLM
top-$k$ prior baseline reaches F1=0.248, indicating that the prior is
predictive on its own, while the LLM pipeline adds further gains.

Our contributions are:
(i) a sparse condition--medication population prior for EHR medication set
prediction;
(ii) a controlled retrieval-source swap inside PACE-RAG with the downstream
pipeline held fixed; and
(iii) sensitivity and stratified analyses showing when population-prior
retrieval is most helpful.

\section{Method}
\label{sec:method}

We formalize the task and the structured population prior, then describe how it
replaces patient-level retrieval inside PACE-RAG.

\subsection{Task formulation}
\label{sec:task}

For each admission $i$, we observe a set of current diagnoses
$D_i=\{d_{i1},\ldots,d_{im}\}$, available prior-admission history $H_i$
including recent medication history when present, and a target medication set
$Y_i\subseteq\mathcal{A}$, where $\mathcal{A}$ is a fixed medication-class
vocabulary. The goal is to predict $\widehat{Y}_i\subseteq\mathcal{A}$ from the
structured admission context. Predictions are evaluated as medication-class
sets, so both missing true medication classes and over-generating irrelevant
classes reduce performance.

\subsection{Structured population prior}
\label{sec:population_prior}

The population prior is a sparse condition--medication memory estimated only
from training admissions under the patient-level split. For each training
admission $i$, we construct a diagnosis set $D_i$ and an observed
medication-class set $Y_i$. For each diagnosis $d$ and medication class $a$, we
compute diagnosis support, pair support, and empirical conditional probability:
\[
  n(d)=|\{i:d\in D_i\}|,\qquad
  n(d,a)=|\{i:d\in D_i,\ a\in Y_i\}|,
\]
\begin{equation}
  \widehat{p}(a \mid d)=\frac{n(d,a)}{n(d)}.
  \label{eq:conditional}
\end{equation}
We also compute the marginal medication frequency
\[
  \widehat{p}(a)=\frac{|\{i:a\in Y_i\}|}{N_{\mathrm{train}}},
\]
where $N_{\mathrm{train}}$ is the number of training admissions, and use it to
estimate base-rate lift,
\[
  \mathrm{lift}(d,a)=\frac{\widehat{p}(a\mid d)}{\widehat{p}(a)}.
\]
The memory stores diagnosis support, diagnosis--medication pair support,
empirical conditional probability, and lift. Low-support pairs are discarded to
reduce noise from rare co-occurrences, and for each diagnosis we retain the
highest-ranked medication classes according to conditional probability and lift.

\subsection{Inference-time aggregation}
\label{sec:aggregation}

At inference time, the population prior does not retrieve patient records. For
a target admission with current diagnosis set $D$, we aggregate diagnosis-level
statistics into a medication score:
\begin{equation}
  s(a\mid D)=
  \frac{1}{|D|}
  \sum_{d\in D}
  \widehat{p}(a\mid d)\mathbf{1}[(d,a)\in\mathcal{I}],
  \label{eq:aggregate_score}
\end{equation}
where $\mathcal{I}$ is the retained condition--medication index after support
filtering. To reduce the dominance of very common medication classes, we also
compute a lift-normalized score over retained medication candidates:
\[
  \ell(a\mid D)=
  \frac{s(a\mid D)}{\widehat{p}(a)}.
\]
This is the multi-diagnosis analogue of $\mathrm{lift}(d,a)$: it replaces the
single-diagnosis conditional probability $\widehat{p}(a\mid d)$ with the
diagnosis-aggregated score $s(a\mid D)$ and normalizes by the marginal
frequency $\widehat{p}(a)$. Since candidates are drawn from retained
training-set condition--medication pairs, $\widehat{p}(a)>0$ for all retrieved
classes. We rank retained medication candidates by $\ell(a\mid D)$ and pass the
top $k$ candidates to the downstream pipeline.

The retrieved context is then a ranked list of medication-class candidates with
coarse probability and lift information. It contains no nearest-neighbor
admissions, patient narratives, or raw training records, so the retrieved
object is directly aligned with the fixed medication vocabulary used for
evaluation.

\subsection{Drop-in retrieval swap inside PACE-RAG}
\label{sec:retrieval_swap}

PACE-RAG retrieves similar patient admissions using FAISS-style vector search
\citep{johnson2019faiss} and passes the retrieved records to a multi-stage LLM
pipeline \citep{huh2026pacerag}. We replace this retrieval call with the
ranked candidates produced by the population prior. These candidates are
formatted into the same context schema expected by the downstream PACE-RAG
stages, while concept extraction, treatment summarization, recommendation
generation, output parsing, and evaluation remain unchanged.
\section{Why Population Priors for EHR RAG?}
\label{sec:why_memory}

Patient-level retrieval and population-prior retrieval expose different
evidence to the LLM. PACE-RAG retrieves a small set of individual admissions
judged similar by an embedding index; the LLM must then infer prescribing
patterns from those examples and map them to the fixed medication vocabulary.
In contrast, the population prior retrieves aggregate condition--medication
statistics computed over all relevant training admissions and already expressed
in the medication-label space.

This distinction is useful for fixed-vocabulary EHR set prediction. Medication
classes often reflect broad prescribing regularities associated with diagnoses
and care contexts, while nearest-neighbor sets can be small, idiosyncratic, or
sensitive to embedding quality. The population prior pools signal across the
training table and is easier to audit, since each retrieved candidate can be
traced to support, conditional probability, and lift. The goal is not to replace
clinical reasoning with counts, but to test whether such statistics provide a
better retrieval interface for the downstream LLM pipeline.
\begin{figure}[t]
  \centering
  \includegraphics[width=\columnwidth]{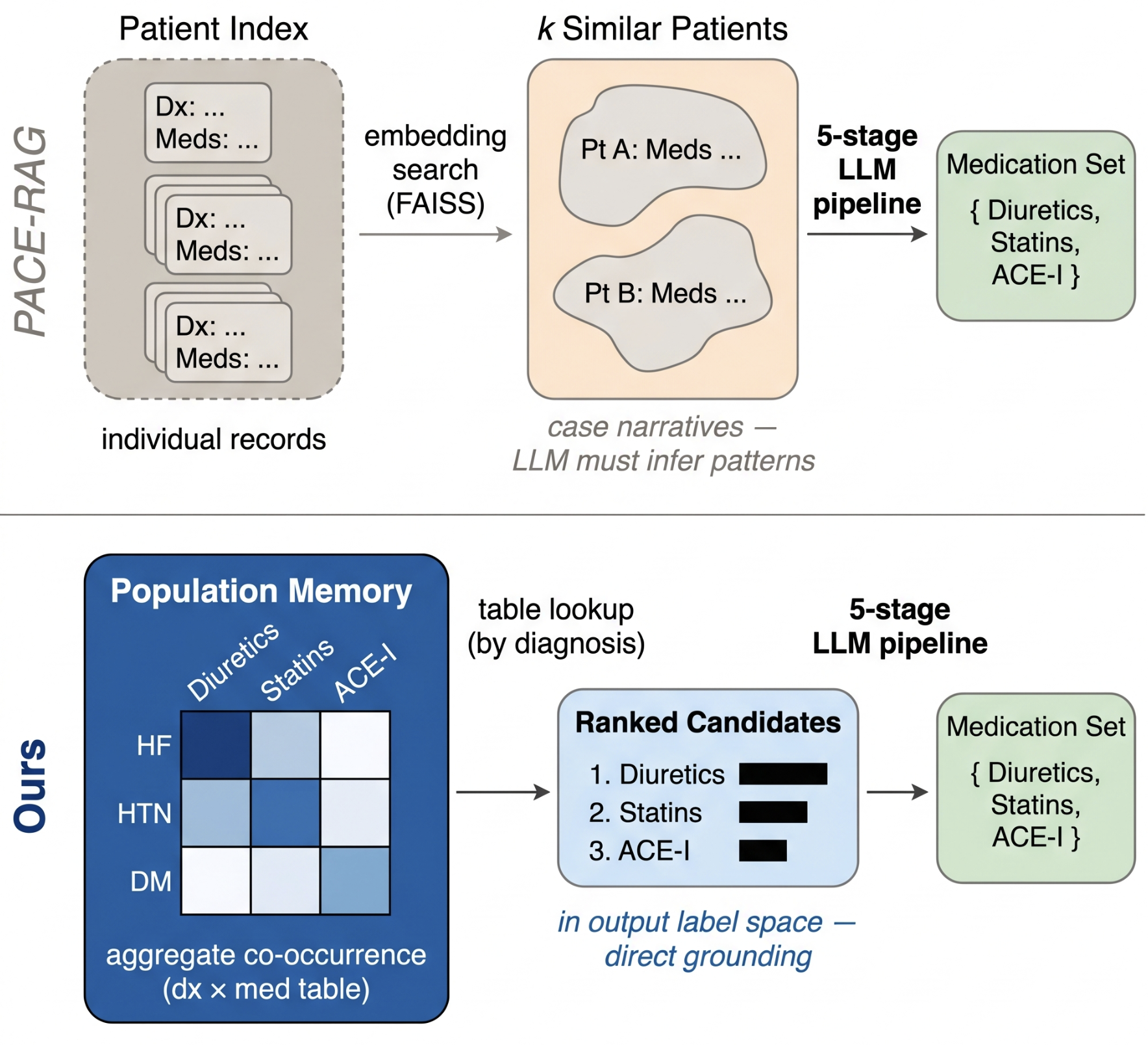}
  \caption{
    Retrieval contrast. Both methods index the same training records
    (diagnoses + medications). PACE-RAG stores individual patient records
    and retrieves $k$ similar cases via FAISS embedding search. Our method
    aggregates these into a condition--medication co-occurrence table; a
    lookup by current diagnoses yields ranked medication-class candidates,
    which replace the FAISS-retrieved records as input to the PACE-RAG pipeline.
  }
  \label{fig:retrieval-swap}
  \vspace{-0.1in}
\end{figure}
\section{Experiments}
\label{sec:experiments}

\paragraph{Data and prediction task.}
We evaluate on a processed MIMIC-IV medication recommendation benchmark
\citep{johnson2023mimic}. Each example corresponds to an inpatient admission
with current diagnosis strings, observed medication classes, and a
patient identifier used to construct chronological prior-admission history. The
split is patient-level: admissions from patients used to build the population
prior are excluded from the test set. The held-out test set contains 1{,}004
admissions. We evaluate predictions as medication-class sets using admission-averaged F1 and
Jaccard similarity. This task should be interpreted as medication-class
prediction from structured EHR context.

\paragraph{Compared systems.}
We choose PACE-RAG as the comparison target because it is a recent
RAG-based medication recommendation system reporting state-of-the-art results
on MIMIC-IV, and its public implementation exposes a patient-level FAISS
retrieval module that can be replaced while keeping downstream LLM stages fixed
\citep{huh2026pacerag}. We compare four configurations. \textbf{Zero-shot GPT-4o-mini} receives only
the current diagnoses and medication vocabulary, with no external retrieval.
\textbf{Population prior, no LLM} directly predicts the top-ranked medication
classes from the structured population prior, providing a diagnostic baseline
for the signal contained in the aggregate table alone. \textbf{PACE-RAG}
\citep{huh2026pacerag} retrieves similar patient admissions using FAISS and
passes the retrieved records through its multi-stage LLM pipeline. We rerun the
public PACE-RAG implementation on the same processed split used by our variant,
rather than copying published numbers. \textbf{PACE-RAG + population prior} is our controlled swap: the FAISS
retrieval call is replaced with a lookup into the condition--medication
population prior, while all downstream PACE-RAG stages remain unchanged.

\paragraph{Controlled retrieval-source swap.}
The main comparison is PACE-RAG versus PACE-RAG + population prior under the
same GPT-4o-mini backbone, temperature, medication vocabulary, output parsing,
and evaluation script. This evaluates whether aggregate condition--medication
statistics are a more useful retrieval object than nearest-neighbor patient
records for fixed-vocabulary EHR set prediction.
\section{Results}
\label{sec:results}

\begin{table}[t]
  \caption{Main results on 1{,}004 held-out MIMIC-IV admissions. PACE-RAG and
  our variant use the same GPT-4o-mini backbone; only the retrieval source differs.
  95\% CIs are computed via bootstrap (2{,}000 resamples).}
  \label{tab:main}
  \centering
  \scriptsize
  \setlength{\tabcolsep}{3pt}
  \begin{tabular}{p{2.8cm} p{3.2cm} cc}
    \toprule
    \textbf{Method} & \textbf{Retrieval source} &
    \textbf{F1 ± 95\% CI} & \textbf{Jac.} \\
    \midrule
    Zero-shot (GPT-4o-mini)
      & none
      & 0.223 $\pm$ 0.026 & 0.137 \\
    Population prior
      & population prior only
      & 0.248 $\pm$ 0.027 & 0.148 \\
    PACE-RAG (GPT-4o-mini)
      & FAISS patient records
      & 0.303 $\pm$ 0.018 & 0.218 \\
    \textbf{PACE-RAG + prior}
      & \textbf{population prior}
      & \textbf{0.338 $\pm$ 0.017} & \textbf{0.237} \\
    \bottomrule
  \end{tabular}
\end{table}

Table~\ref{tab:main} shows that the retrieval source substantially affects
PACE-RAG performance. Replacing FAISS patient-record retrieval with the
structured population prior improves F1 from 0.303 to 0.338, a +0.035 absolute
gain (+11.5\% relative), and improves Jaccard from 0.218 to 0.237. Since both
systems share the same downstream LLM pipeline and evaluation procedure, the
gain suggests that the form of retrieved evidence---aggregate medication
candidates rather than neighboring patient records---is a key determinant of
performance.

The remaining rows help separate three sources of performance: the base LLM,
the retrieved statistics, and the interaction between retrieval and the
multi-stage LLM pipeline. Zero-shot GPT-4o-mini reaches F1=0.223, indicating
that diagnosis names and the medication vocabulary alone provide a weak but
nontrivial signal. The population-prior-only baseline improves to F1=0.248,
showing that the aggregate condition--medication table captures prescribing
regularities that are not recovered by prompting the LLM without retrieval.

The full PACE-RAG + population prior system reaches F1=0.338, above both the
zero-shot and prior-only baselines. This suggests that the LLM is not merely
copying the highest-ranked prior candidates, but using them as structured
evidence that can be filtered and reconciled with the admission context. The
comparison to FAISS-based PACE-RAG further suggests that already-aggregated
medication evidence is a better retrieval interface for this task than a small
set of patient narratives.

\paragraph{Diagnostic complexity.}
The improvement is not driven by a single aggregate effect. As shown in
Table~\ref{tab:diagnostic}, the population prior outperforms FAISS patient
retrieval in every diagnostic-complexity bin. The largest absolute gain occurs
for admissions with one or two current diagnoses (+0.054 F1), where
nearest-neighbor retrieval may be especially brittle because the LLM must infer
a prescribing pattern from only a few retrieved cases.

The gain remains positive for admissions with 10 or more diagnoses (+0.029 F1),
where nearest-neighbor retrieval might be expected to be more informative. This
suggests that aggregate condition-level statistics remain useful even when the
diagnostic context is richer, possibly by providing a stable candidate set that
the LLM can filter against the broader admission context. Inference-time latency
measurements are reported in Appendix~\ref{app:efficiency}.
\begin{table}[t]
  \caption{Performance stratified by diagnostic complexity, measured by the
  number of current diagnoses $|D|$. The population prior improves over
  FAISS patient retrieval in every bin.}
  \label{tab:diagnostic}
  \centering
  \small
  \begin{tabular}{lccc}
    \toprule
    \textbf{$|D|$ bin} & \textbf{$N$} & \textbf{PACE-RAG F1} &
    \textbf{+Pop Prior F1} \\
    \midrule
    1--2  & 43   & 0.184 & \textbf{0.238} \\
    3--5  & 158  & 0.192 & \textbf{0.238} \\
    6--9  & 223  & 0.280 & \textbf{0.318} \\
    10+   & 580  & 0.351 & \textbf{0.380} \\
    \midrule
    All   & 1004 & 0.303 & \textbf{0.338} \\
    \bottomrule
  \end{tabular}
\end{table}

\paragraph{Candidate-count sensitivity.}
We selected $k$ on a held-out validation set of 1{,}000 training patients
(prior rebuilt without them) and obtained $k=30$ as the best value
(val~F1=0.427). Test-set F1 is essentially flat from $k=5$ onward
(range 0.338--0.340), confirming the result is not sensitive to this
hyperparameter; full details are in Appendix~\ref{app:topk}.
\section{Discussion and Limitations}
\label{sec:discussion}

For fixed-vocabulary EHR set prediction, the retrieval object supplied to a
clinical RAG pipeline matters. The population-prior variant supports the central
structured-data hypothesis of the paper: aggregate condition--medication
statistics can provide a more task-aligned retrieval signal than a small set of
nearest-neighbor admissions.

The result also highlights a practical advantage of tabular population memory.
Retrieval is a lookup into a sparse count table whose entries can be inspected
through support, conditional probability, and lift, without an inference-time
embedding model, vector index, or similarity search. This makes the retrieved
evidence more auditable and easier to maintain while still allowing the
downstream LLM pipeline to synthesize and filter candidate medication classes.

\paragraph{Limitations.}
This study should be interpreted as medication-class prediction, not prescribing
advice: we do not model dose, route, timing, contraindications, drug--drug
interactions, or patient-specific safety constraints. Population priors also
reflect historical prescribing patterns, which may vary across institutions,
time periods, and patient subgroups. External validation, recalibration,
fairness analysis, and prospective clinical evaluation would be required before
any deployment consideration.
\bibliography{example_paper}
\bibliographystyle{icml2026}

\newpage
\appendix
\onecolumn

\section{Population Prior Implementation Details}
\label{app:population_memory}

The population prior is a non-learned retrieval object built only from training
admissions under the patient-level split. It has no trainable weights, no
embedding model, and no vector index, so the controlled swap with FAISS isolates
the \emph{form} of retrieved evidence: aggregate condition--medication
statistics versus patient records.

For each diagnosis string, the prior stores retained medication classes with
diagnosis support, pair support, empirical conditional probability, and lift.
Low-support condition--medication pairs below a fixed minimum count threshold
are removed before ranking. At inference time, matching pairs for the current
diagnosis set are aggregated as described in Section~\ref{sec:aggregation},
ranked by the lift-normalized score, and the top $k$ medication classes are
returned. The retrieved context passed to the LLM contains only medication-class
names with coarse probability or lift tiers; it does not include raw training
records or nearest-neighbor patient narratives.
\section{PACE-RAG Rerun and Controlled Retrieval Swap}
\label{app:pacerag}

We rerun the public PACE-RAG implementation
(\url{https://github.com/ChaeYoungHuh/PACE-RAG}) on the same processed
MIMIC-IV tables, patient-level split, medication vocabulary, and evaluation
script used for the population-prior variant. We do not copy the originally
reported PACE-RAG numbers, because the published implementation uses different
LLM backbones (Llama-3.1-8B / Qwen3-8B) and its reported cohort statistics
differ from those produced by our preprocessing. Our preprocessing of the same
public codebase yielded 163{,}594 training patients, 342 unique ATC medication
classes in the training pool, 202 in the test set, and 1{,}004 test
admissions, compared to 196{,}110 patients, 406 training-pool classes, 254
test-set classes, and 1{,}094 test admissions in the original paper. In this
paper, all LLM-based systems are run with GPT-4o-mini at temperature 0.0.

The controlled swap keeps the PACE-RAG pipeline fixed after retrieval. In the
baseline, PACE-RAG retrieves similar patient admissions using FAISS and passes
those retrieved records to the downstream LLM stages. In our variant, the FAISS
retrieval call is replaced by a lookup into the structured population prior.
The population-prior candidates are formatted into the same context schema
expected by the downstream PACE-RAG stages.

Thus, all post-retrieval stages remain unchanged, and the main comparison is a
retrieval-source ablation rather than a comparison between different LLM
pipelines. Both our PACE-RAG baseline and PACE-RAG + population prior operate
on the same processed split, backbone, medication vocabulary, and evaluation
script.

\section{Candidate-Count Sensitivity}
\label{app:topk}

We swept $k \in \{3,5,10,20,30\}$ candidates on a held-out validation set
of 1{,}000 patients sampled by patient id from the training data.
The population prior was rebuilt from the remaining training patients so that
validation admissions are never in the prior used to evaluate them.
Table~\ref{tab:topk} reports validation F1 used for model selection and test
F1 for robustness analysis. Validation F1 increases monotonically with $k$, so
we select $k=30$. On the test set, F1 is essentially flat from $k=5$ onward
(range 0.338--0.340), indicating that the result is not sensitive to the exact
candidate count once a small candidate set is available. The main result uses
$k=30$, the validation-selected value.

\begin{table}[h]
  \caption{Candidate-count sensitivity. $k$ is selected on the validation set;
  test F1 is reported only to assess robustness.}
  \label{tab:topk}
  \centering
  \small
  \begin{tabular}{rcc}
    \toprule
    \textbf{$k$} & \textbf{Val F1} & \textbf{Test F1} \\
    \midrule
     3 & 0.319 & 0.330 \\
     5 & 0.337 & 0.338 \\
    10 & 0.367 & 0.340 \\
    20 & 0.406 & 0.339 \\
    \textbf{30} & \textbf{0.427} & 0.338 \\
    \bottomrule
  \end{tabular}
\end{table}

\clearpage
\section{Inference-Time Efficiency}
\label{app:efficiency}

\begin{table}[h]
  \caption{Per-admission inference latency, measured as the median over 10
  sequential calls with GPT-4o-mini at temperature 0.0. Both systems run the
  identical five-stage LLM pipeline; the only difference is the retrieval
  source.}
  \label{tab:efficiency}
  \centering
  \small
  \begin{tabular}{lc}
    \toprule
    \textbf{Method} & \textbf{Latency (median)} \\
    \midrule
    PACE-RAG (FAISS patient retrieval) & 10.2 s \\
    \textbf{PACE-RAG + population prior} & \textbf{8.5 s} \\
    \bottomrule
  \end{tabular}
\end{table}

End-to-end latency is dominated by the same sequential LLM calls in both
systems. The population prior therefore does not add LLM cost and slightly
reduces retrieval overhead in this implementation. More importantly, it removes
the FAISS index and embedding model from the inference path. At serving time,
retrieval reduces to a lookup in a compact condition--medication table rather
than maintaining a vector store and running embedding-based nearest-neighbor
search.

\end{document}